# Face Detection Using Improved Faster RCNN

Changzheng Zhang, Xiang Xu, Dandan Tu*
Huawei Cloud BU, China
{zhangzhangzheng, xuxiang12, tudandan}@huawei.com

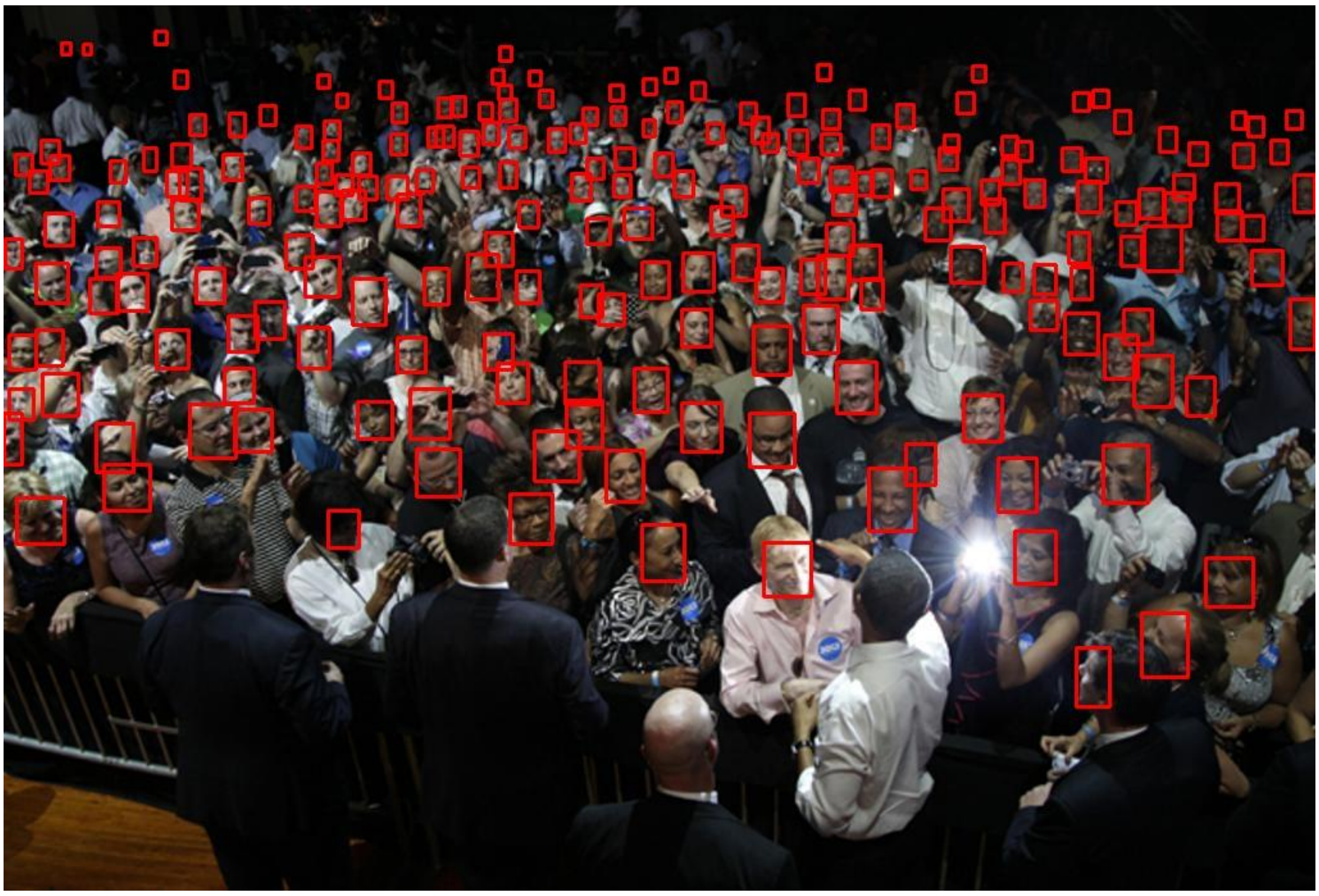

Figure1.Face detection results of FDNet1.0

## Abstract

Faster RCNN has achieved great success for generic object detection including PASCAL object detection and MS COCO object detection. In this report, we propose a detailed designed Faster RCNN method named FDNet1.0 for face detection. Several techniques were employed including multi-scale training, multi-scale testing, light-designed RCNN, some tricks for inference and a vote-based ensemble method. Our method achieves two 1th places and one 2nd place in three tasks over WIDER FACE validation dataset (easy set, medium set, hard set).

## 1 Introduction

Face detection is a critical step for many face-related applications, such as face alignment, face verification, face identification, crowed behavior analysis etc. However, small size, occlusion, illumination, pose deformation, expression and other disadvantageous factors often appear in real-world images [9], which bring great challenges to face detection. Recently, generic object detection based on deep convolution neural networks (CNNs) has achieved great success. It utilizes modern object detectors including one stage methods (e.g., YOLO[1-2],SSD[3-4]) and two stage methods(e.g., Faster RCNN[5-6],RFCN[7-8]). One stage methods refer broadly to architectures that use a single feed-forward full convolutional neural network to directly predict each proposal's class and corresponding bounding box without requiring a second stage per-proposal classification operation and box refinement . Two stage methods, especially Faster RCNN achieves better performance than one stage methods over many object detection benchmarks [10-11]. In the Faster R-CNN setting, object detection happens over two pipes. In the first pipe, input image is directly processed by a feature extractor (e.g., Vgg16[27], Inception[28-31], ResNet101[6]) without any hand engineering, and features at the selected intermediate layer (e.g.,“conv5_3”[27],“res4f”[6]) will be fed to a convolutional layer, which simultaneously predict objectiveness scores and region bounds at each location on a regular grid according to predefined stride. The first pipe is also called region proposal network (RPN). In the second pipe, these proposals with higher scores in the RPN are used to crop features from the same intermediate feature map which are subsequently fed to the remainder of the feature extractor (e.g., two full connected layer[27], 5th block[6]) in order to predict a class and class-specific box refinement for each proposal.

*Corresponding author

Face detection [9, 32] has achieved great success thanks to the appearance of one stage method and two stage methods. However, there are still some issues with these methods that can be improved with elaborate design of the details. In this report, we propose a detailed design Faster RCNN method named FDNet1.0 for face detection, which achieves more decent performance than previous methods [13-25]. A deformable layer with fewer channels is attached to the backbone network to produce a “thin” feature map, which is subsequently fed to a full connected layer, building an efficient yet accurate two-stage detector [12]. At testing time, we also find a comparable mean average precision (mAP) be achieved when the top-ranked proposals (e.g., 6000) are directly selected [13] without NMS in the RPN stage over WIDER FACE dataset. It is also beneficial for hard set to keep the small proposals (<16 pixels width/height) at training and testing stage as there are many tinny faces of WIDER FACE dataset. Furthermore, the multi-scale training and testing strategy are also applied in our work.

Our key contributions are summarized as follows:

(1)A light head based two-stage framework named FDNet1.0 is developed for face detection.

(2)Some useful tricks are found to improve final face detection performance including multi-scale training, multi-scale testing, keep the small proposals at training and testing stage, directly select top-ranked proposals (e.g., 6000) without NMS in the RPN stage for R-CNN, a vote-based NMS ensemble strategy.

(3)Our framework achieves two 1st places and one 2nd place in three tasks over WIDER FACE validation dataset (easy, medium, hard), one illustrative example of our results in the crowd case can be found in Figure 1.

## 2 Related Work

Face detection is one of the most fundamental and challenging problems in computer vision, and has been extensively studied for decades. Compared against these hand-engineered features, a lot of progress for face detection has been made in recent years due to utilizing of modern object detectors, including Faster R-CNN, R-FCN, SSD, YOLO and their extensions.

**Hand-engineered approaches:**

A cascaded AdaBoost face detector [33] is proposed to detect face by using Haar-like features. Based on this groundbreaking work, more advanced hand-engineered features and more powerful machine learning algorithms [34-36] are developed to improve face detection performance. Additionally, deformable part models (DPM) is also employed for face detection by several research groups, which achieve remarkable performance [37-39].

**Single-stage approaches:**

CascadeCNN [14] proposes a strategy to detect face coarse to fine. A mutli-task learning method [15] named MTCNN is present to predict face and landmark location simultaneously. Dense-Box [16] employs a fully deep convolutional neural network to directly predict face confidence and corresponding bounding box. UnitBox [17] introduces a novel intersection-over-union (IoU) loss to predict bounding box, which regresses the four bounds of a predicted box as a whole unit. SAFD [18] and RSA unit [19] devote to handle scale explicitly using CNN or RNN. $S^3$FD [20] presents a single shot scale-invariant face detector which achieves good result on WIDER FACE datasets. Very recently, FAN [21] presents an effective face detector based on feature pyramid network, which obtains state-of-the-art results.

**Two-stage approaches:**

Face R-CNN [22] employs a new multi-task loss function based on Faster R-CNN framework to enhance performance. CMS-RCNN [23] is proposed to enhance face detect performance by exploiting contextual information. Convnet [24] introduces an end-to-end multi-task discriminative learning framework to increase occlusion robustness. Based on R-FCN [7], Face R-FCN [25] re-weights embedding responses on score maps and eliminates the effect of non-uniformed contribution in each facial part using a position-sensitive average pooling.

# 3 Proposed Approach

Faster RCNN, with two fully connected layers or all the convolution layers in ResNet 5-th stage to predict RoI classification and regression, consumes a large memory and computing resource. RFCN is fully convolutional with almost all computation shared on the entire image, but it has poor performance compared to Faster RCNN. Inspired by [12], we develop a light-head Faster RCNN for face detection with good performance and inference speed. In this section, we will present our method in detail.

### 3.1 Light-Head Faster RCNN

Based on Faster RCNN, we make several effective modifications for improving detection performance. The architecture of our framework is depicted in Figure 2.

ResNet architecture plays the role of feature extractor, the "thin" feature maps is built by a deformable layer before Region-of-Interest (RoI) warping, which will exploit image context and be robust to variations. And a single fully-connected layer is used in the R-CNN subnet.

We use ResNet-v1-101 as backbone network to extract high-level feature. The stride of ResNet-v1-101 is fixed to 16 pixels, which is not good enough for detecting different scale faces. Therefore a large kernel-based deformable layer is attached on backbone network to exploit image context and be robust to variations, where "D" stand for deformable. The output channel size of the deformable layer is 512, and each ROI will be resized to 512×7×7, which will be fed to the following fully connected layer with 2048 channels.

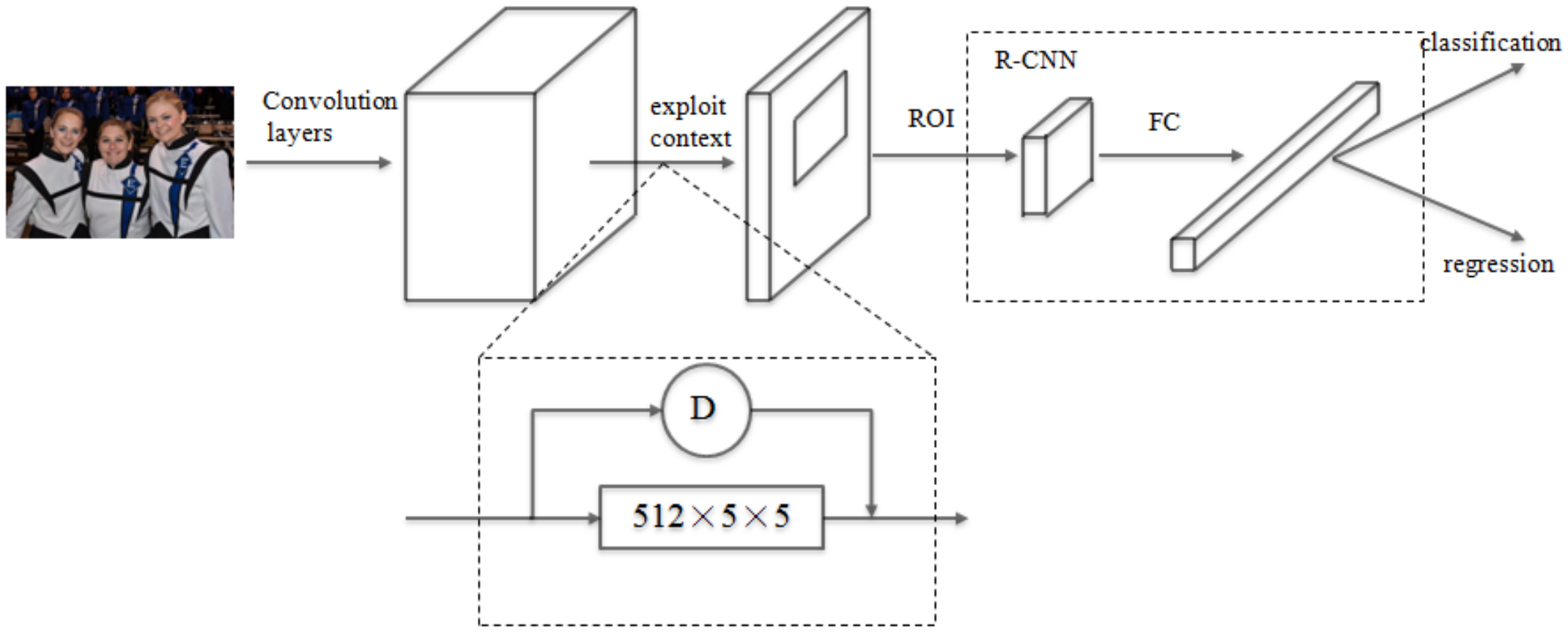

Figure2. Overview of our approach. A deformable layer with 512 channels was employed to build "thin" feature maps with exploiting image context simultaneously before RoI warping. Hence, RCNN can be designed with light-head to improve inference speed.

Additionally, anchors are carefully designed to obtain better location samples. The aspect ratio is set to 1, 1.5, 2 and the scale is set to $16^2$, $32^2$, $64^2$, $128^2$, $256^2$, $512^2$ based on statistical analysis on the training dataset. These smaller anchors are very helpful for sufficiently capturing tiny face. As WIDER FACE dataset contain many extremely tinny faces (< 16 pixels width/height), we keep these small proposals (< 16 pixels width/height) valid in the training and testing time [13]. The experiments show that our method can achieve better performance.

### 3.2 Multi-Scale Training and Testing

The trained model can also be robust on different scale faces when both multi-scale training and testing strategy are used. In our method, the shorter side is resized to 600,800,1000,1200,1400 pixels according to the statistical analysis on the training dataset. Unlike [21], we only use horizontal image flipping augmentation, and no other hard example mining method is used in the training stage. In the testing phase, the shorter side of each image is also resized to 600,800,1000,1200,1400 pixels and tested independently. Then all of the output results are merged. Next, a voted-based NMS strategy is adapted. We firstly delete the output bounding box whose IOU is lower than 0.3 with any other bounding boxes to suppress false positive samples, and then NMS is used to get the best bounding boxes.

# 4 Experiments

We perform evaluation on WIDER FACE dataset [9] which contains more challenges, including small scale, illumination, occlusion, background clutter and extreme poses when compared with other benchmarks. A total of 393,703 labeled faces in 32,203 images from 61 different scenes are collected, of which 40% are chosen as train set, 10% as validation set and other 50% as test set. The validation set and the testing set are also divided into easy set, medium set and hard set according to the detection difficulties. It is noticed that our method is trained only on the train set and evaluate on both validation set and test set. Better performance might be achieved by merging the train set and the validation set for training. More detailed results of WIDER FACE are shown in Figure 4.

## 4.1 Implementation Details

Single NVIDIA Tesla K80 is used for training and testing. Mini batch size is set to 1 considering memory consumption. Specifically, ResNet_v1_101 trained on ImageNet-128w is used for Faster RCNN feature extraction. It is helpful to freeze the first two blocks in the training stage as data size of WIDER FACE is not so large. A deformable layer is used to output a "thin" feature map with exploiting image context. Aspect ratios (1, 1.5, 2) and scales ($16^2$, $32^2$, $64^2$, $128^2$, $256^2$, $512^2$) are carefully designed to capture better locations of faces in the RPN stage, and the number of filters for the RPN layer is set as 512. The anchors with highest IoU score or IoU score with the ground truth above 0.7 are defined as positive. The anchors whose IoU score with the ground truth that is lower than 0.3 are defined as negative, while whose IoU score above 0.3 but lower than 0.7 will be ignored. The similar settings of anchors are used in the R-CNN stage. The anchors with IoU score with the ground truth above 0.5 are assigned as positive, IoU score that is lower than 0.3 is defined as negative, IoU score above 0.3 but lower than 0.5 will be ignored. By the way, the batch size of RPN and R-CNN is respectively assigned as 256 and 128. The initial learning rate is set to 1e-3, and decrease to 1e-4 after 20w iterations. Weight decay is and momentum is set to 1e-4 and 0.9 respectively.
In testing stage, multi-scale testing strategy is adapted to be robust to different scale faces. Specifically, the shorter side of each image is also resized to 600, 800, 1000, 1200, 1400 pixels and tested independently. And a voted-based NMS strategy is used to get the final result. We also find top-ranked 6000 proposals are directly selected without NMS during testing can boost 0.1%, 0.3% and 0.6% on easy set, medium set and hard set respectively.

## 4.2 Comparison on Benchmarks

Our model is trained on the train set and evaluated on WIDER FACE validation set. Compared with the recently published top approaches, FDNet1.0 wins two 1st places (easy set = 95.9%, medium set = 94.5%) and one 2nd place (hard set = 87.9%) on the validation set, as illustrated in Figure 3. We believe that more kinds of data augmentation and hard example mining [26] would further boost detection performance.

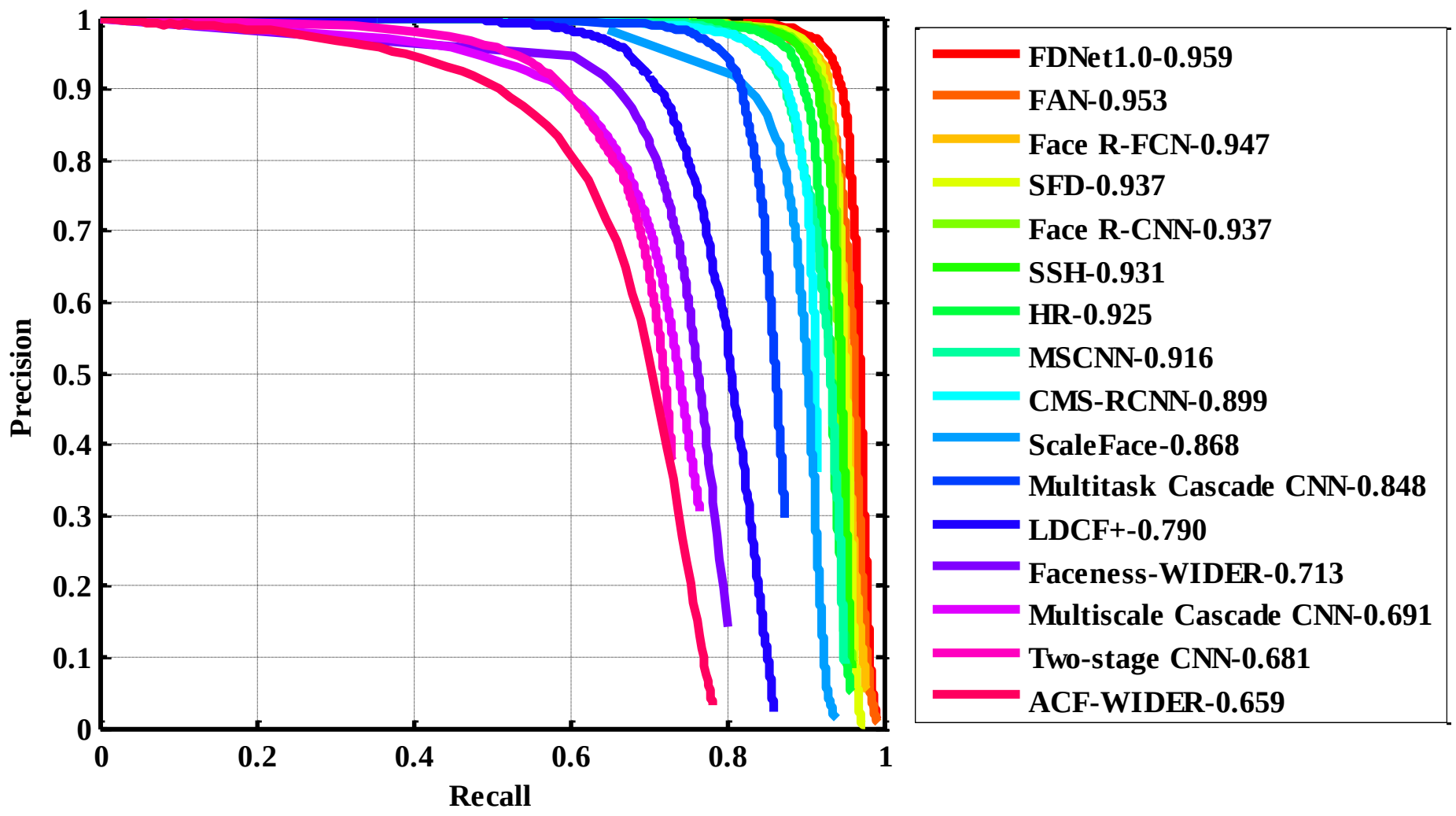

(a) Easy set: validation

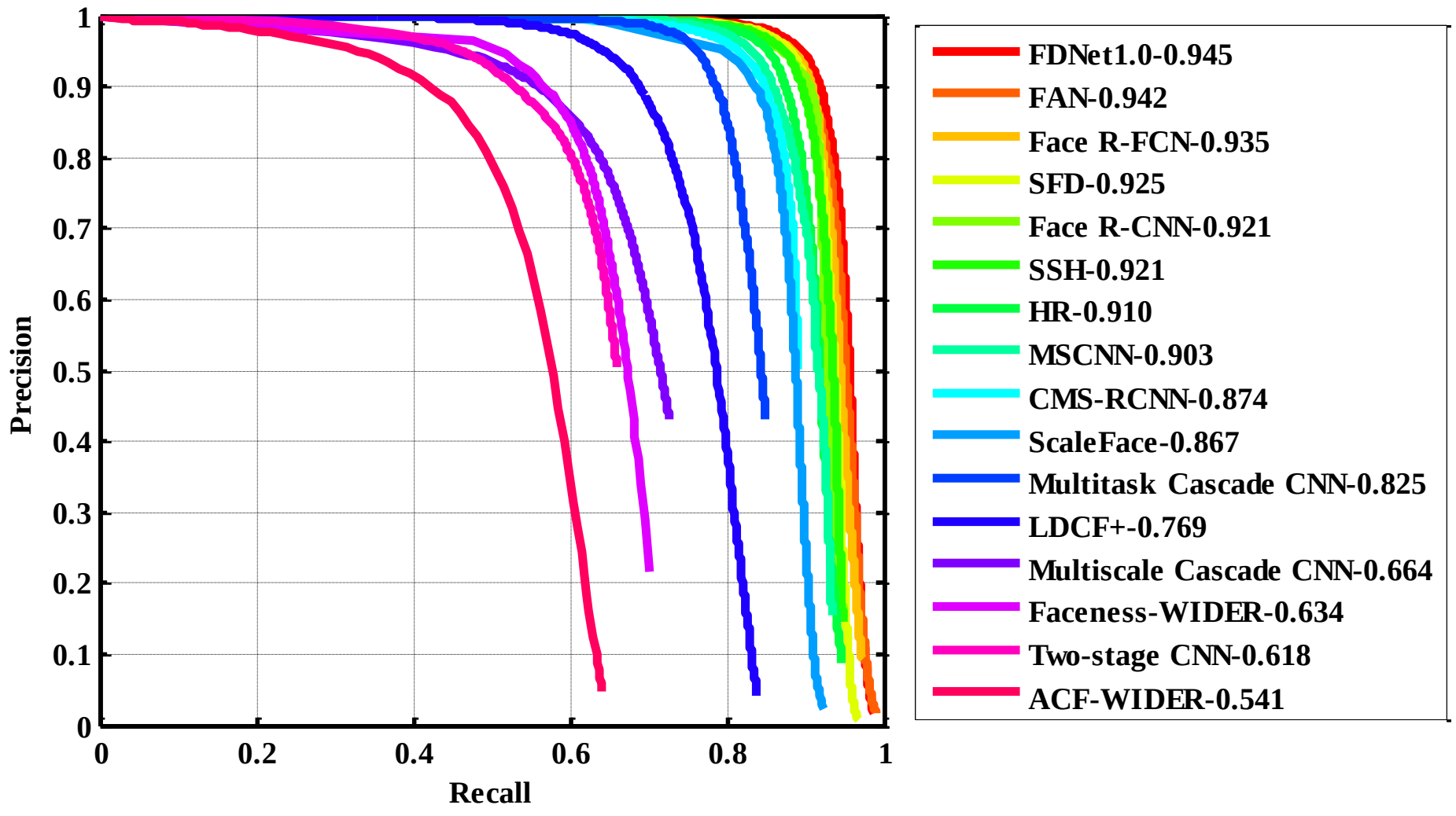

(b) Medium set: validation

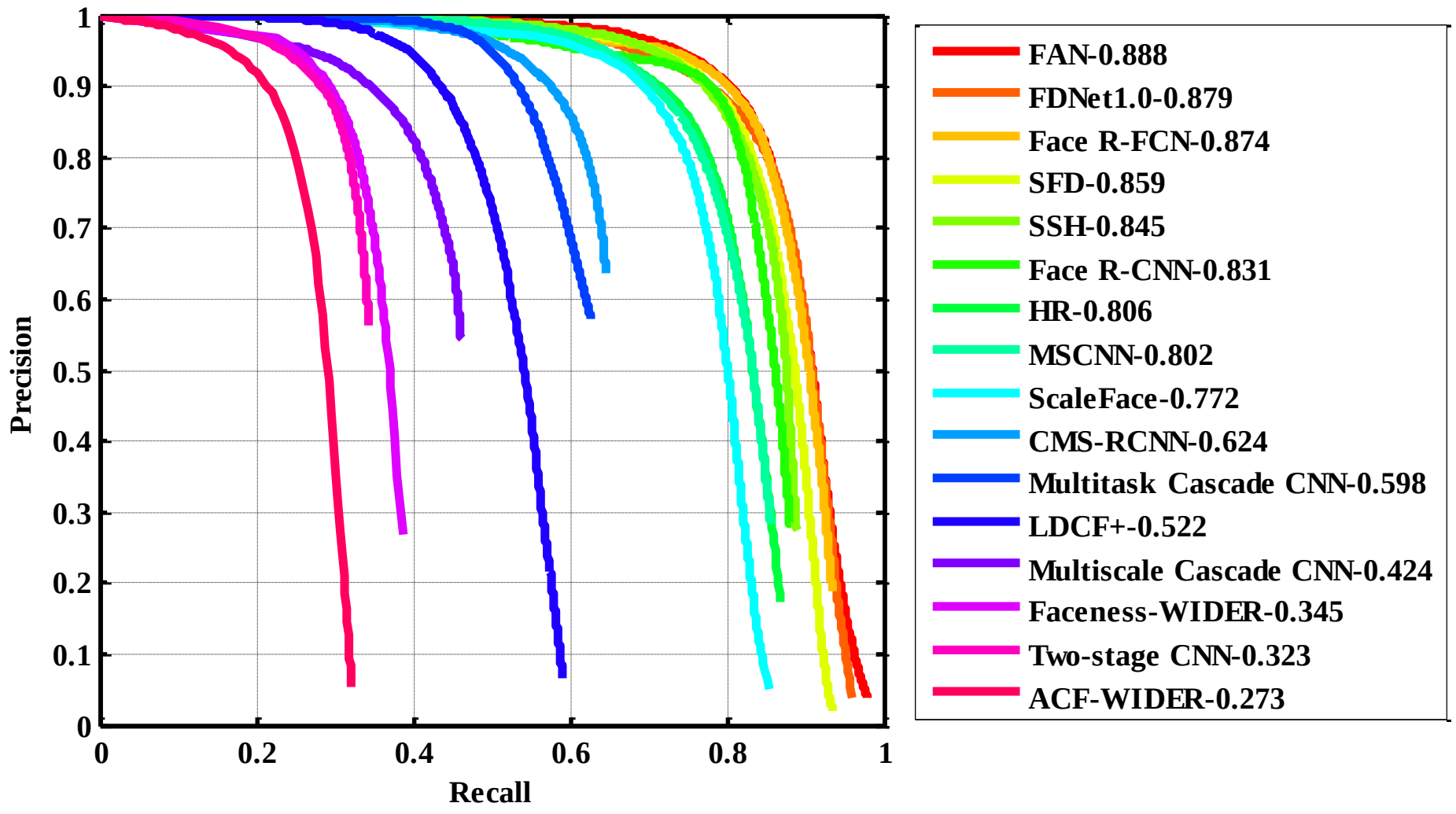

(c) Hard set: validation

Figure3. Precision-Recall curves on WIDER FACE's validation easy set, medium set and hard set.

# 5. Conclusion

In this paper, we propose a novel framework named FDNet1.0 for face detection. FDNet1.0 improves Faster RCNN by integrating several efficient techniques for better performance. Experimental results on challenging WIDER FACE dataset validate the effectiveness of our proposed algorithm. In the future, we will try more kinds of data augmentation and hard example mining which may further boost detection performance. We will also consider some ideas for faster inference speed, e.g. designing light backbone.

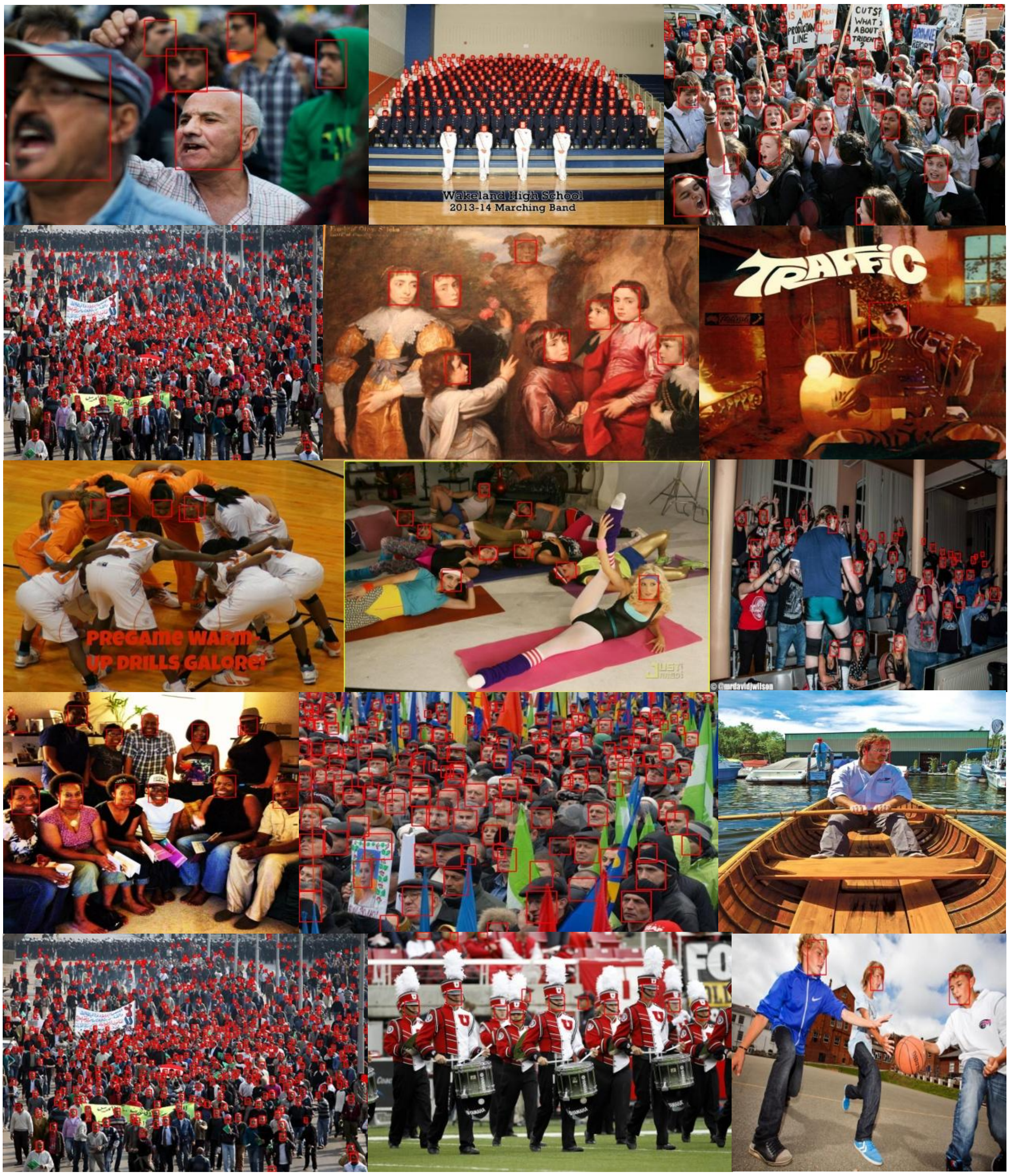

Figure.4. Examples of our detected results on the WIDER FACE validation set, including small size, occlusion, illumination, pose deformation, expression etc.